\documentclass{article}

 \usepackage[position, final]{neurips_2026}

\usepackage[utf8]{inputenc} %
\usepackage[T1]{fontenc}    %

\usepackage{url}            %
\usepackage{booktabs}       %
\usepackage{amsfonts}       %
\usepackage{nicefrac}       %
\usepackage{microtype}      %
\usepackage{xcolor}         %
\usepackage{graphicx}
\usepackage{epigraph}
\usepackage{wrapfig}

\definecolor{citecolor}{rgb}{0.3,0.3,0.3}
\definecolor{linkcolor}{rgb}{0,0,0.5}
\usepackage[pagebackref,colorlinks=true,citecolor=citecolor,linkcolor=linkcolor]{hyperref}

\usepackage{amsmath} 

\newcommand{\E}[2]{\mathbb{E}_{#1}{\left[#2\right]}}

\newcommand{\M}{\mathcal{M}}
\renewcommand{\S}{\mathcal{S}}
\newcommand{\A}{\mathcal{A}}
\newcommand{\D}{\mathcal{D}}
\newcommand{\supp}{\mathrm{supp}}

\newcommand{\defeq}{\mathrel{\mathop:}=}

\title{From Static Policies to Adaptive Priors \\in Offline Reinforcement Learning}

\author{%
Tianwei Ni\textsuperscript{1,2}\thanks{
Now at Google DeepMind. Correspondence to: \texttt{twni2016@gmail.com}. \\Blog post: \href{https://twni2016.github.io/blogs/policyprior/main.html}{twni2016.github.io/blogs/policyprior/main.html}.
}\quad  
Vineet Jain\textsuperscript{1,3}\quad  
Akash Karthikeyan\textsuperscript{1,2}\quad  
Pierre-Luc Bacon\textsuperscript{1,2} \\
$^{1}$Mila - Quebec AI Institute\quad  
$^{2}$Université de Montréal\quad  
$^{3}$McGill University
}

\begin{document}

\maketitle

\begin{abstract}

Offline reinforcement learning (RL) has traditionally focused on learning policies for direct deployment under conservative objectives, where uncertainty outside the offline dataset is treated pessimistically to ensure robustness. We argue that this formulation becomes incomplete when an offline-trained policy is subsequently updated through online interaction, as increasingly occurs in modern intelligent systems through test-time adaptation and online fine-tuning. This position paper argues that, in such settings, the objective of offline RL should extend beyond immediate deployment and instead prioritize learning \textit{adaptive policy priors}: policies that preserve the capacity to improve during subsequent interaction through memory, exploration, and self-correction. We formalize this perspective as \textit{adaptive offline reinforcement learning} (AORL), distinguish it from offline-to-online RL, and explain why adaptability becomes important under distributional shift, limited dataset coverage, and changing test-time conditions. We further discuss Bayesian offline RL as one principled direction for constructing adaptive policy priors by preserving epistemic uncertainty over plausible environments. Finally, we outline connections, open challenges, and research directions for treating offline RL as preparation for future experience rather than as a static deployment problem.

\end{abstract}

\section{Introduction}

\setlength{\epigraphwidth}{0.6\textwidth}
\epigraph{
The excess of history has seized the plastic force of life.}
{Friedrich Nietzsche, \textit{On the Use and Abuse of History for Life}}
\vspace{-0.5em}

Offline reinforcement learning (RL) studies how to optimize a policy using a static dataset of trajectories, without relying on online interaction during training\footnote{We refer to ``training'' as parameter updates in the policy, in contrast to in-context learning that does not modify parameters~\citep{brown2020language}.}. In its classical formulation~\citep{levine2020offline}, the objective is to learn a policy for \textit{direct deployment}: the offline-learned policy is treated as a static, final decision rule whose performance is expected to derive entirely from offline data rather than from subsequent interaction. This deployment-oriented formulation makes distributional shift particularly challenging~\citep{kumar2019stabilizing}: actions chosen outside the dataset may lead to states where value estimates are unreliable, errors compound over time, and correcting such errors through further interaction is not part of the learning objective.

As a result, the dominant design principle of offline RL has been \textit{conservatism}: uncertainty about out-of-dataset actions is treated pessimistically so that learned policies remain safe and robust under limited support. This principle appears through imitation objectives~\citep{pomerleau1988alvinn,fujimoto2021minimalist}, policy constraints~\citep{fujimoto2019off,wu2019behavior,peng2019advantage}, pessimistic value estimation~\citep{kumar2020conservative,an2021uncertainty,kostrikov2021offline}, or uncertainty penalties~\citep{yu2020mopo,kidambi2020morel,bai2022pessimistic}.
This conservative formulation has driven much of the progress in offline RL~\citep{reed2022generalist,lee2022multi,kumar2023offline,park2025horizon}, particularly in benchmark settings where policies are evaluated under fixed environment conditions.\looseness=-1 

A fundamental limitation of this formulation is that it leaves little room for improvement after offline learning, even though such improvement is especially important in the \textit{era of experience}~\citep{silver2025welcome}. In this era, performance depends not only on knowledge extracted from static datasets, but also on the ability to acquire and exploit new information through online interaction in continual and open-ended environments~\citep{hadsell2020embracing,khetarpal2020towards,hughes2024position,dupoux2026ai}. In these settings, offline optimization is not the end of learning; rather, it produces an initial policy whose quality should also be judged by how effectively it improves after deployment.\looseness=-1

Recent practice already reflects this shift across domains, including large language models (LLMs) and robotics, where offline-learned policies are routinely followed by mechanisms that learn from subsequent experience, such as test-time in-context learning~\citep{brown2020language,wei2022chain}, planning over imagined trajectories~\citep{yao2023tree,snell2025scaling}, or online RL fine-tuning~\citep{guo2025deepseek,guo2025improving}. Under this broader role, excessive conservatism carries an opportunity cost: actions assigned near-zero probability during offline learning may become difficult to recover later, even when subsequent interaction would reveal them to be beneficial.

\begin{figure}[t]
    \centering
    \includegraphics[width=\linewidth]{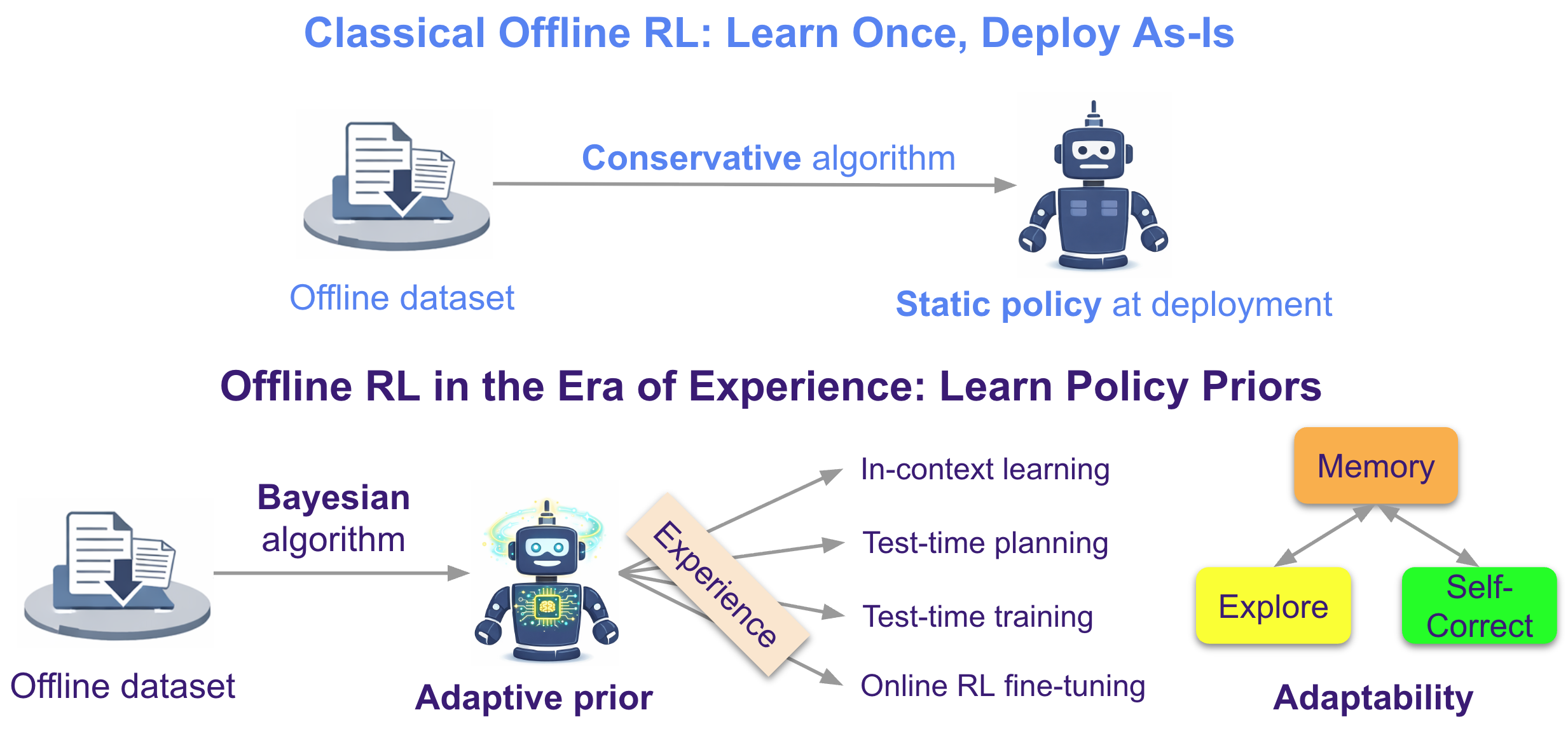}
    \vspace{-1em}
    \caption{Top: \textbf{Classical offline RL} learns a static policy from a static dataset under conservative objectives. Bottom: \textbf{Adaptive offline RL} learns a policy prior that remains improvable through subsequent experience, including in-context learning, planning, or online fine-tuning. Here, \textbf{adaptability} arises from the interaction between memory, exploration, and self-correction. Bayesian perspective is highlighted as one principled direction for constructing such adaptive priors.}
    \vspace{-1em}
    \label{fig:summary}
\end{figure}

In this position paper, we use the term \textit{adaptive offline reinforcement learning} (AORL) to refer to settings where offline RL should prioritize adaptability beyond direct deployment. \textbf{Our position is that when policies will continue improving through interaction, offline RL should learn adaptive policy priors rather than static policies.} This differs from \textit{offline-to-online RL}~\citep{nair2020awac}, where offline learning primarily serves as initialization for subsequent online \textit{training}. In AORL, the objective of the offline stage itself is to preserve the capacity for later adaptation, even when online interaction is limited or occurs without parameter updates.

Under this view, offline learning should not eliminate uncertainty solely for immediate robustness, but preserve sufficient behavioral flexibility for later adaptation. Such adaptability requires three ingredients: \textit{memory}, so that decisions depend on online history; \textit{exploration}, so that uncertain but potentially valuable actions remain reachable; and \textit{self-correction}, so that new evidence can revise early mistakes during interaction. Actions outside the dataset are therefore not inherently undesirable; rather, they reflect epistemic uncertainty that can later be resolved through in-context learning, planning, or fine-tuning. \autoref{fig:summary} summarizes this shift from static policies to adaptive policy priors.\looseness=-1

One principled direction for AORL arises from Bayesian perspectives~\citep{ghosh2022offline,ni2026long}. In this view, offline RL is cast as an epistemic POMDP~\citep{ghosh2021generalization}: limited dataset coverage induces a posterior distribution over plausible MDPs that agree on observed data while differing beyond dataset support. The resulting optimal policy is naturally \textit{adaptive in context}: by conditioning on online history, it can first explore uncertain but potentially good actions, then exploit the most promising ones. Bayesian offline learning therefore offers a concrete interpretation of adaptive policy priors whose value emerges through test-time interaction rather than as static behavior at deployment.\looseness=-1

The remainder of this paper is organized as follows. We first formalize adaptive offline reinforcement learning and define adaptive policy priors through memory, exploration, and self-correction. We then explain why these properties become increasingly important when offline learning is followed by test-time adaptation or online fine-tuning, especially under distributional shift, limited high-quality coverage, and changing deployment conditions. Next, we present Bayesian offline RL as a principled direction that naturally realizes adaptive behavior under epistemic uncertainty, while discussing alternative views that continue to prioritize conservatism or treat adaptation purely as an online RL problem. Finally, we outline broader connections, open challenges, and research directions for establishing adaptive offline RL as a practical learning paradigm.

\section{Formulation of Adaptive Offline Reinforcement Learning}
\label{sec:formulation}

\subsection{Offline RL Problem}

We consider the standard offline RL setting~\citep{levine2020offline} for a discrete-time, infinite-horizon, discounted-reward MDP defined by the tuple $\mathcal{M}^* = (\mathcal{S}, \mathcal{A}, \rho^*, P^*, R^*, \gamma)$.\footnote{The same setting extends naturally to partially observable MDPs (POMDPs)~\citep{kaelbling1998planning}, where the policy observes $o_t$ emitted from state $s_t$, and states in the history are replaced by observations.}
Here, the state space $\mathcal{S}$, the action space $\mathcal{A}$, and the discount factor $\gamma \in (0,1)$ are assumed known, while the environment components, including the initial state distribution $\rho^* \in \Delta(\mathcal{S})$, the transition function $P^*: \mathcal{S} \times \mathcal{A} \to \Delta(\mathcal{S})$, and reward function $R^*: \mathcal{S} \times \mathcal{A} \to \Delta(\mathbb R)$, are \textit{unknown}. 

Instead of interacting with $\mathcal{M}^*$ during training, the offline learner receives a static offline dataset $\mathcal{D} = \{\tau^i\}_{i=1}^N$ of trajectories collected by an unknown behavior policy $\beta$. Each trajectory $\tau = (s_0, a_0, r_1, s_1, a_1, r_2, \dots)$ is generated by
$$
s_0 \sim \rho^*, \quad
a_t \sim \beta(h_t), \quad
s_{t+1} \sim P^*(s_t, a_t),\quad
r_{t+1} \sim R^*(s_t, a_t),\quad 
\forall t \ge 0,
$$
where the behavior policy may depend on the interaction history. We define the \textit{history} at time $t$ as
$$
h_t \defeq (s_0, a_0, r_1, s_1, \dots, a_{t-1}, r_t, s_t),
$$
with $h_t \in \mathcal{H}_t$ and $\mathcal{H}_t$ denoting the corresponding history space.\footnote{For applications where rewards are unavailable at test time, histories are defined without past rewards.} Histories evolve recursively as $h_{t+1} = h_t \oplus (a_t, r_{t+1}, s_{t+1})$ for all $t \ge 0$, with $h_0 \defeq (s_0)$ and $\oplus$ denoting sequence concatenation. In practice, trajectories are finite due to truncation by time limits, while infinite-horizon objectives are recovered through value bootstrapping.

The \textit{ideal goal} for offline RL is to find a policy that maximizes expected discounted return under the true environment. In the most general form, this policy may depend on interaction history, denoted as $\pi : \mathcal{H}_t \to \Delta(\mathcal{A})$. 
\begin{equation}
\max_{\pi}\, J(\pi; \mathcal M^*) \defeq \E{\tau}{\sum_{t=1}^\infty \gamma^{t-1} r_t \,\middle|\, \pi, \mathcal M^*}.
\end{equation}

Because $\mathcal M^*$ is inaccessible during offline optimization, this objective cannot be optimized directly. The key difficulty is epistemic uncertainty in state-action regions poorly covered by $\D$. We define the \textbf{behavioral support} of a state $s$ within $\D$ as 
$$
\supp_\D(s) = \{a \mid \Pr\nolimits_\D(a\mid s) > \epsilon\} \subseteq \A,
$$ 
where $\Pr_{\mathcal D}(\cdot \mid s)$ denotes the empirical action distribution in $\mathcal D$ and $\epsilon > 0$ is a small threshold. For states $s$ not observed in the dataset,  $\supp_\D(s)=\emptyset$. Epistemic uncertainty over the dynamics $P^*(s,a)$ and reward $R^*(s,a)$ therefore remains substantial over actions $a \in \mathcal A \setminus \supp_\D(s)$ for each $s\in \S$. 
Different treatments of this uncertainty give rise to different algorithmic principles, including conservative offline RL~\citep{levine2020offline}, Bayesian offline RL~\citep{ghosh2022offline}, and optimistic offline RL~\citep{agarwal2020optimistic}.

\textbf{Offline training and online testing.}
Although offline RL (including AORL) excludes online data collection during training, the learned policy is still evaluated through online interaction with the environment at test time. AORL focuses on test-time adaptation during this interaction, involving no parameter changes or only temporary ones.

\subsection{Core Components of Adaptive Policy Priors}

\begin{table}[h]
    \centering
    \caption{Comparison between classical offline RL and adaptive offline RL.}
    \resizebox{0.9\textwidth}{!}{%
    \begin{tabular}{l|cc}
    \textbf{Property} & \textbf{Classical Offline RL} & \textbf{Adaptive Offline RL} \\
    \midrule
    Decision rule & Typically Markovian & History-dependent \\
    Uncertainty treatment & Within offline support & Preserve flexibility \\
    Design focus & Offline stage & Offline + test-time stages \\
    Core capabilities & Safety, robustness & Memory, exploration, self-correction \\
    \end{tabular}
    }
    \vspace{-0.5em}
    \label{tab:formulation}
\end{table}

\textbf{History dependence in decision making (memory).}
Classical offline RL typically assumes a Markovian decision rule for MDPs, where actions are selected solely from the current state through a policy $\pi: \mathcal S \to \Delta(\mathcal A)$~\citep{levine2020offline}. Adaptive offline RL (AORL) instead advocates history-dependent decision making even in fully observable MDPs, because limited offline coverage leaves uncertainty about environment dynamics and rewards. This dependence can be implemented \textit{explicitly} through a history-dependent policy $\pi: \mathcal H_t \to \Delta(\mathcal A)$~\citep{chen2021offline}, or \textit{implicitly} through online adaptation of latent variables~\citep{ghosh2022offline,liu2023design}, plans~\citep{janner2022planning}, or even policy parameters~\citep{xu2025test}. In this sense, history dependence serves as memory: the same state $s$ may require different actions when preceding observations imply different beliefs about the environment. Memory is therefore the basic mechanism that allows online interaction to influence future decisions.

\textbf{Behavioral flexibility (exploration and self-correction).}
Classical offline RL typically restricts learned actions to the behavioral support $\supp_\D(s)$ or heavily penalizes actions outside it~\citep{levine2020offline}. In AORL, by contrast, out-of-support actions are not treated as intrinsically undesirable: some may be potentially useful but remain unresolved under offline data alone because of epistemic uncertainty rather than evidence of poor value. Preserving non-negligible probability on such actions allows the policy to explore uncertain behaviors during interaction. Equally importantly, because these actions may fail, the policy should be able to self-correct by using online feedback to revise future decisions rather than committing to early mistakes~\citep{wang2024theoretical,kumar2025training}. Exploration and self-correction therefore form two \textit{complementary} roles of behavioral flexibility, both of which rely on memory.

\textbf{Co-design of offline optimization and test-time adaptation.}
Classical offline RL primarily designs algorithms within the offline optimization stage, treating test-time interaction as passive deployment~\citep{levine2020offline}. In AORL, by contrast, offline optimization and test-time adaptation are jointly considered: the offline-learned policy is evaluated by how well it supports memory, exploration, and self-correction under the true environment.

These components define the basic requirements for an \textbf{adaptive policy prior}: the policy should retain memory from interaction history while preserving sufficient flexibility to explore uncertain behaviors and correct them online. \autoref{tab:formulation} summarizes the key differences between classical and adaptive offline RL. These requirements appear naturally in several forms of adaptation that occur after offline optimization, which we describe next.

\subsection{Forms of Online Adaptation}

Adaptation occurs after offline optimization, when an offline-learned policy is allowed to interact with the true environment $\M^*$ and improve with the resulting online interaction data, referred to as \textit{experience}~\citep{silver2025welcome}. These adaptation mechanisms differ in whether policy parameters are updated and how much online data they require. Common forms are summarized in \autoref{tab:adaptation}. These forms are not mutually exclusive and can be composed in practice.

\begin{table}[h]
    \centering
    \caption{Common forms of adaptation after offline optimization.}
    \resizebox{0.9\textwidth}{!}{%
    \begin{tabular}{l|ccc}
      \textbf{Form of Adaptation}  & \textbf{Online Data} & \textbf{Parameter Update?} & \textbf{Mechanism}   \\
      \midrule
     Test-time in-context learning    & Limited & No  & Implicit inference  \\
     Test-time planning & Limited & No  & Explicit inference  \\ 
     Test-time training & Limited & Yes  & Temporary update \\ 
     Online RL fine-tuning & Extended & Yes  & Persistent update 
    \end{tabular}
    }
    \vspace{-0.5em}
    \label{tab:adaptation}
\end{table}

\textbf{Test-time in-context learning.} 
The simplest form of adaptation occurs when decision making is history-dependent, so that behavior changes directly through interaction history without modifying policy parameters or performing additional optimization. This mechanism is commonly referred to as \textit{in-context learning}~\citep{brown2020language,moeini2025survey}, and has often been interpreted as implicit Bayesian inference over latent tasks~\citep{xie2022explanation}. It differs from \textit{in-weight learning}, where adaptation occurs through parameter updates.
A special case of in-context learning is \textit{self-improvement}~\citep{shinn2023reflexion}, where explicit reward signals may be unavailable at test time, yet the agent can still improve through informative observations that reduce uncertainty about the environment. Self-improvement includes settings where the uncertainty lies in transition dynamics rather than reward function~\citep{packer2018assessing}, as well as settings where feedback is conveyed through language observations~\citep{cheng2024llfbench,klissarov2026improving}. 

\textbf{Test-time planning.}
In-context learning relies primarily on the generalization ability of the offline-learned policy, which may be insufficient when the true environment $\M^*$ differs substantially from what is specified by $\D$. Test-time planning addresses this limitation by improving policy decisions with online search over future trajectories, often using learned world models~\citep{argenson2020model,zhou2025diffusion,wei2025plangenllms}. Recent planning approaches further enable trajectory optimization through probabilistic inference (e.g., diffusion-based denoising) without requiring an explicit world model~\citep{janner2022planning,ajay2023is}.

\textbf{Test-time training.}
Adaptation may also occur through parameter updates using a limited amount of online data collected at test time~\citep{finn2017model,sun2020test}. Unlike in-context learning or planning, test-time training modifies the policy itself, often through maximizing Q-value on test-time states without reward signals~\citep{park2024value,xu2025test}. By iteratively updating parameters on online trajectories, test-time training encodes experience as parametric memory.
Although test-time training lies outside the classical offline RL formulation, it is relevant under AORL because the offline-learned policy is viewed as a prior that should support efficient improvement from limited online experience.

\textbf{Online RL fine-tuning.}
A more extended form of adaptation is online RL fine-tuning, where the offline-learned policy is subsequently improved through persistent online optimization. This includes \textbf{offline-to-online RL}~\citep{nair2020awac,lee2022offline} and RL post-training of foundation models~\citep{guo2025deepseek}. Offline-to-online RL describes a particular adaptation pipeline, whereas AORL concerns the objective of the offline stage. Thus, an offline-to-online method instantiates AORL \textit{when} its offline stage is explicitly designed to enhance test-time adaptability, rather than merely provide a strong initialization.

\section{Why Offline RL Needs Adaptive Policy Priors?}
\label{sec:why}

We distinguish two sources of uncertainty that motivate adaptive policy priors. First, incomplete offline data coverage leaves parts of the environment unidentified; online distributional shift and exploration bottlenecks expose the consequences of this uncertainty during deployment. Second, the deployment environment itself may change after offline data collection.

\textbf{Incomplete offline coverage and its deployment consequences.}
Classical offline RL effectively bounds policy improvement by the quality of behaviors represented in the dataset~\citep{jin2021pessimism,uehara2021pessimistic}. This limitation becomes pronounced when high-quality actions are absent or underrepresented~\citep{ni2026long}. In practice, collecting high-quality trajectories is expensive, and in open-ended domains even carefully curated datasets may remain incomplete. This aligns with the broader observation that human-generated data alone imposes a ceiling on improvement~\citep{silver2025welcome}. 

This coverage gap appears through online distributional shift. Even when the deployment environment is unchanged, a learned policy may encounter states that are rare or absent in the offline dataset. Small errors can compound over time, producing \textit{covariate shift} as in imitation learning~\citep{ross2010efficient}, and this remains a practical bottleneck in offline RL~\citep{park2024value}. Since such OOD states are rarely trained explicitly under conservative objectives, the learned policy may become brittle when uncertainty matters most. Adaptability helps address this by preparing the offline-learned policy to use memory and online feedback to \textbf{self-correct} once OOD states are encountered. In this sense, adaptability does not conflict with safety: adaptive policies can be designed to remain risk-averse while still using online evidence to revise decisions under uncertainty~\citep{rigter2023one}.

\begin{figure}[h]
    \centering
    \includegraphics[width=0.7\linewidth]{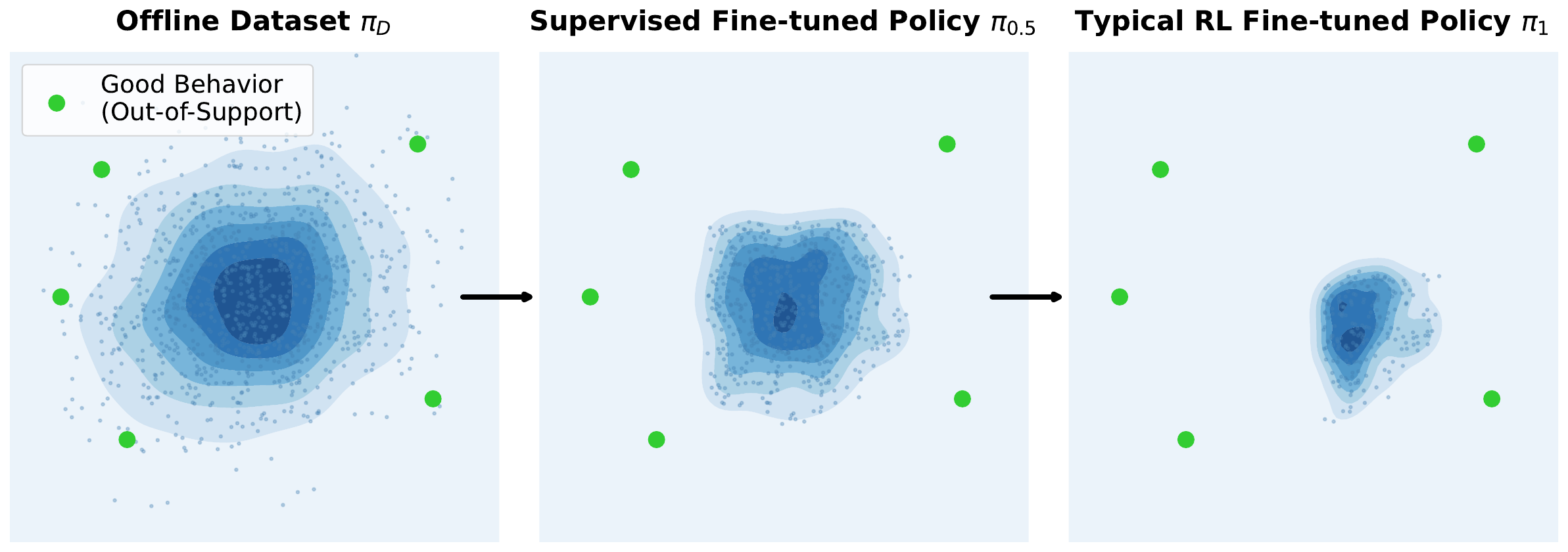}
    \vspace{-0.5em}
    \caption{\small\textbf{Schematic support contraction in a continuous-action bandit.} Each point is a possible two-dimensional action; green points are potentially good out-of-support actions. Blue shading denotes the density of offline-stored or learned behaviors. Support may shrink across SFT and online RL fine-tuning, making underrepresented good actions harder to recover~\citep{yue2025does}.}
    \vspace{-0.5em}
    \label{fig:support_contraction}
\end{figure}

Continuing with online RL does not automatically resolve the problem, because exploration
from a narrow static policy can be slow~\citep{luo2023finetuning,zhao2023enoto}. In addition, recent evidence in foundation model post-training further suggests that RL often mainly reweights behaviors already present in the policy prior rather than reliably discovering new ones~\citep{yue2025does,zhao2025echo}. 
As a result, behavioral support may contract further during fine-tuning, as illustrated in \autoref{fig:support_contraction}, making underrepresented but potentially good actions increasingly difficult to recover. Although stronger exploration strategies or prolonged fine-tuning can alleviate this effect~\citep{liu2025prorl}, a complementary and often cheaper solution is to prepare an \textbf{exploratory} policy prior that already preserves diverse modes beyond well-supported actions in the offline dataset.

\textbf{Test-time condition change.}
The preceding arguments assume that deployment occurs in the same MDP that generated the
offline data. In realistic open-ended domains, dynamics, rewards, or task structure may shift after offline data collection, so actions that were optimal under the offline dataset may no longer remain optimal at test time. Prior work has shown that offline-learned policies can become brittle under such mismatches, including settings where variation is specified through environment parameters~\citep{liang2023adaptdiffuser}, external data sources~\citep{lyu2024odrl}, natural-language instructions~\citep{karthikeyan2025genplan}, or left unspecified~\citep{mediratta2024gengap}. Adaptability helps address this limitation by enabling policies to use online interaction to infer changed conditions, explore alternative behaviors, and self-correct decisions as new evidence accumulates.

\vspace{-0.5em}
\section{A Bayesian Perspective for Adaptive Policy Priors}
\label{sec:bayesian}

\begin{figure}[h]
    \centering
    \includegraphics[width=0.7\linewidth]{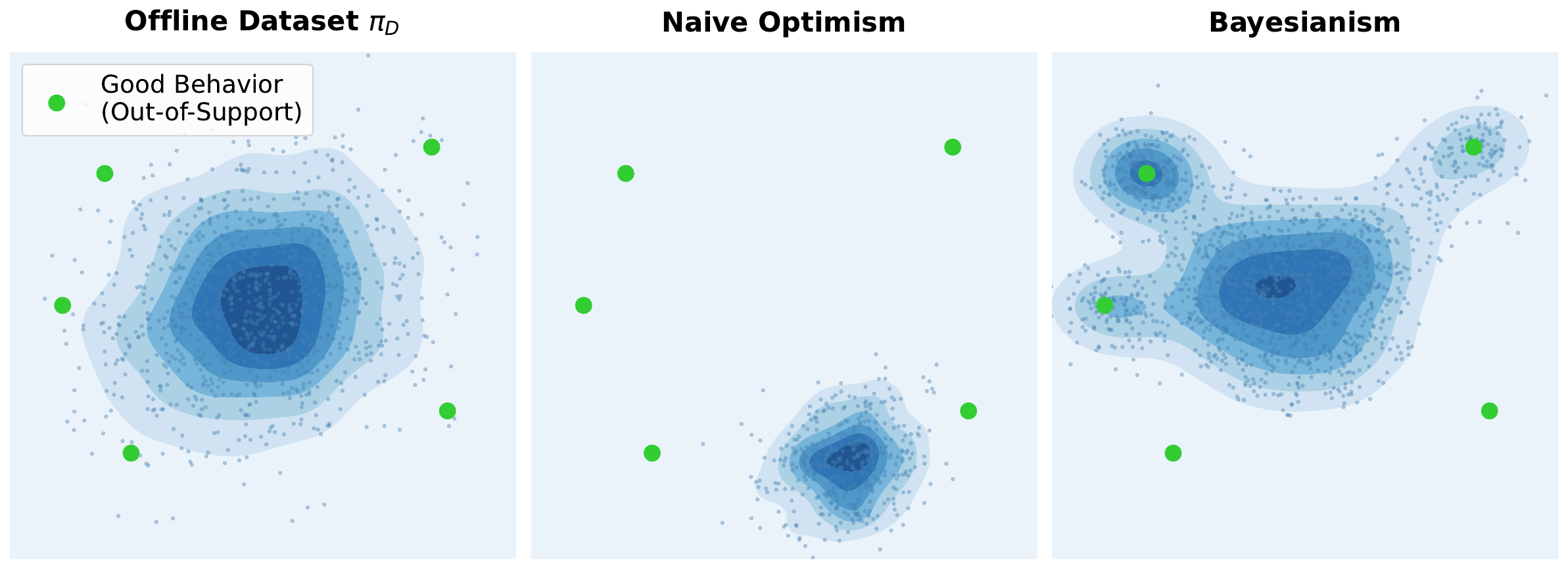}
    \vspace{-0.5em}
    \caption{\small\textbf{Schematic illustration of uncertainty treatment in a continuous-action bandit.} Using the same legend as \autoref{fig:support_contraction}, Bayesian offline RL treats unseen actions as uncertain rather than inherently bad or reliably good. In contrast, naive optimism in model-free off-policy RL may assign unreliable values to out-of-support actions due to extrapolation error~\citep{fujimoto2019off}.}
    \vspace{-0.5em}
    \label{fig:bayesian}
\end{figure}

A natural question is whether adaptive policy priors admit a principled computational interpretation. One promising direction arises from Bayesian principles in offline RL~\citep{ghosh2022offline}, building on Bayes-adaptive MDPs~\citep{duff2002optimal}, whose practical potential has recently been demonstrated~\citep{chen2021offline,choi2024diversification,ni2026long}. Rather than collapsing uncertainty conservatively or treating unseen actions optimistically, Bayesian offline RL maintains multiple plausible environment hypotheses consistent with the offline dataset, as illustrated in \autoref{fig:bayesian}. Prior work provides \textit{concrete} illustrations: \citet[Section~3]{ghosh2022offline} and \citet[Section~3]{ni2026long} show that Bayes-adaptive policies strictly outperform static policies in bandit-like problems.

\textbf{Bayesian formulation of offline RL.}
In Bayesian model-based offline RL, the offline learner maintains a Bayesian world model consistent with the offline dataset $\mathcal D$. A prior distribution $\Pr(\mathcal M)$ over MDPs induces a posterior $\Pr(\mathcal M \mid \mathcal D)$ after observing the dataset, where each MDP $\mathcal M$ is specified by environment parameters $(\rho, P, R)$. The offline RL objective becomes
\begin{equation}
\max_{\pi}\, \E{\M \sim \Pr(\M \mid \D)}{J(\pi; \M)}.
\end{equation}
Because each sampled MDP's identity is unknown to the agent, decision making becomes a partially observable problem over environment hypotheses, often called an \emph{epistemic POMDP}~\citep{ghosh2021generalization,ghosh2022offline}. The optimal policy is therefore naturally history-dependent: it maintains Bellman consistency under the posterior over plausible MDPs. By contrast, model-free offline RL may assign unreliable values to out-of-support actions without an explicit mechanism for Bellman consistency.

\textbf{Connection to adaptive policy priors.}
This formulation yields memory, exploration, and self-correction through an agent-side belief learned during offline policy training. Conditioned on its interaction history $h_t$, the policy implicitly approximates
\begin{equation}
\begin{split}
\underbrace{\Pr(\mathcal M \mid \mathcal D, h_t)}_{\text{Agent's online posterior}} \propto \underbrace{\Pr(\mathcal M\mid \mathcal D)}_{\text{Offline Posterior}} \underbrace{\Pr(h_t \mid \mathcal M, \mathcal D)}_{\text{Online Likelihood}},
\end{split}
\end{equation}
where the offline posterior serves as the effective prior for online interaction, while new experience reweights the belief toward MDPs that better explain the observed history. An uncertain action may have value of information when its outcome distinguishes among plausible MDPs and improves future decisions. A Bayes-adaptive policy therefore \emph{explores} when the expected benefit of resolving uncertainty exceeds the action's expected cost. After observing the outcome, the policy can either reinforce the behavior or \emph{self-correct} by updating its belief and subsequent actions. Adaptability thus arises from continuously refining the MDP posterior through online experience.

\vspace{-0.5em}
\section{Alternative Views}
\label{sec:alternatives}

\textbf{Conservatism as the first principle.}
A dominant alternative view in offline RL is that conservatism should remain the primary design principle, motivated mainly by a technical concern: offline value estimation is prone to extrapolation error and value overestimation~\citep{fujimoto2019off,kumar2019stabilizing}. Conservative objectives therefore restrict policy improvement toward in-distribution behavior to stabilize learning; in high-stakes deployment settings, this also aligns naturally with safety requirements.
Our position differs when an offline-learned policy is expected to continue improving online. In this setting, uncertainty need not be fully eliminated offline, and excessive conservatism may suppress behaviors needed for later adaptation. Recent evidence suggests that value overestimation can also be controlled through alternative mechanisms such as long-horizon planning~\citep{ni2026long}, weakening the case for conservatism as the default first principle.
Importantly, \textit{adaptability does not rule out moderate conservatism}: some degree of conservatism may still be necessary~\citep{wendl2026safe}, and can even be adjusted at test time~\citep{swazinna2023userinteractive}. 
What matters is preserving enough memory and behavioral flexibility for the policy to revise decisions through interaction. The key distinction is whether offline learning aims to produce a static, final policy or an improvable prior.\looseness=-1

\textbf{Adaptation should be left to test-time methods and online RL, not offline RL.} 
Another alternative view is that offline RL need not itself produce an adaptive prior; its role is simply to provide a stable initialization for subsequent test-time adaptation or online RL. Under this perspective, one may even blur the distinction between offline and online learning by directly training online RL from scratch with offline data~\citep{song2023hybrid,ball2023efficient}. This is reasonable in settings where online interaction is abundant, but incomplete when online experience is limited or costly. If offline training collapses the policy too narrowly---for example through behavioral cloning or strongly conservative objectives---test-time methods and online RL may inherit a restricted search space and struggle to recover underrepresented but valuable behaviors~\citep{yue2025does}. Finally, when offline learning is intended to produce \textit{reusable} priors rather than task-specific warm starts, its objective should therefore extend beyond fast initialization to preserving behavioral flexibility for later adaptation.

\textbf{Adaptability already appears in many offline RL methods.} Another alternative view is that offline RL already contains mechanisms associated with adaptation, making an explicit shift in objective unnecessary. Common history-dependent policies trained by offline RL~\citep{chen2021decision,rafailov2021offline} use past context during action generation, yet typically remain constrained by behaviors within the offline dataset. Test-time planning methods~\citep{argenson2020model,janner2022planning} improve decisions through online search, but the candidate trajectories they consider are usually still constrained by offline support. Similarly, \textit{stitching}~\citep{fu2020d4rl} enables compositional generalization from offline data, but this remains a passive recombination of in-distribution behaviors rather than adaptation driven by new online evidence. These examples suggest that adaptive ingredients are increasingly present in offline RL, but common formulations still do not explicitly preserve exploration and self-correction under later interaction.

\vspace{-0.5em}
\section{Open Challenges and Research Directions}
\label{sec:open_directions}

\textbf{Benchmarks and evaluation for adaptability.}
Existing offline RL benchmarks evaluate many important settings, including high-dimensional observations~\citep{gulcehre2020rl,lu2023challenges}, sparse rewards and stitching~\citep{fu2020d4rl,park2025ogbench}, stochasticity~\citep{qin2022neorl}, and safety~\citep{liu2024datasets}. However, they rarely evaluate whether an offline-learned policy can \emph{improve through subsequent interaction}. Prior work on adaptability has so far focused either on bandit-like settings~\citep{ghosh2022offline,ni2026long} or illustrative task modifications~\citep{liang2023adaptdiffuser,zhou2025diffusion,karthikeyan2025genplan}. 
Two notable exceptions are the off-dynamics RL benchmark~\citep{lyu2024odrl} and the generalization benchmark~\citep{mediratta2024gengap}. In \citet{lyu2024odrl}, test-time condition changes are specified through externally provided datasets, whereas in \citet{mediratta2024gengap}, changes are conveyed either through text instructions in WebShop or remain unspecified in Procgen. If AORL is to become a meaningful research direction, progress will require more benchmarks that explicitly measure the capability for test-time adaptation.

On the offline side, dataset design should include at least two cases: \emph{underrepresented but valuable actions}, where useful behaviors appear only sparsely in the data, and \emph{unseen but plausible actions}, where beneficial behaviors are absent altogether and must be discovered through later interaction. On the online side, evaluation should go beyond a single fixed test environment and instead consider \textit{a distribution of environments}: (1) environments fully aligned with the offline data, and (2) environments with test-time condition changes under varying degrees of specification.

More broadly, AORL benchmarks should treat \textit{multi-episode interaction} as the basic unit of evaluation, akin to meta-RL setups~\citep{duan2016rl}. One canonical evaluation objective is
\begin{equation}
J_{\mathrm{eval}}(\pi; P_{\mathrm{test}}, K) = \E{\mathcal M \sim P_{\mathrm{test}}}{\E{}{\sum_{k=1}^K \sum_{t=1}^{T_k} \gamma^{t-1} r_{k,t} \mid \pi, \mathcal M}},
\end{equation}
where $P_{\mathrm{test}}(\mathcal M)$ is the evaluation distribution over deployment environments, $K$ is the number of test-time episodes, and $T_k$ is the horizon of episode $k$. The policy may condition on interaction history accumulated across episodes, allowing experience from earlier episodes to improve later decisions.\looseness=-1

This objective captures both initial performance and subsequent adaptation. Benchmarks should report returns across episodes to measure adaptation speed, while alternative protocols may use different episode weightings or regret-based metrics.

\textbf{Theory of adaptive offline RL.} 
Classical offline RL theory typically relies on coverage assumptions over good actions and Markovian policies. Extending these guarantees to history-dependent policies that can discover unseen but valuable actions through online interaction remains largely open. 
Key questions include which class of offline datasets favors adaptive priors over static policies, when in-context learning suffices versus when parameter updates are necessary, and how guarantees should change under test-time condition shift.

\textbf{Overcoming value overestimation.} 
A major challenge in AORL is that classical \textit{value overestimation} re-emerges once explicit conservatism is relaxed~\citep{fujimoto2019off,kumar2019stabilizing,sims2024edge}. Because adaptive priors must preserve uncertain actions for later exploration, avoiding overestimation \textit{without} collapsing back to conservative behavior becomes a central algorithmic difficulty. Recent evidence from Bayesian offline RL suggests that sufficiently long-horizon rollouts can mitigate this problem~\citep{ni2026long}. This points to a broader direction: scaling long-horizon planning in model-based offline RL, potentially informed by recent progress in long-horizon world modeling~\citep{li2025uncertainty,lin2026admv}.

\textbf{Scalable Bayesian inference.} 
A practical bottleneck of the Bayesian direction in \autoref{sec:bayesian} is that exact posterior inference over MDPs is typically intractable. In practice, uncertainty is often approximated through model ensembles, where disagreement serves as a proxy for epistemic uncertainty~\citep{yu2020mopo}. Improving the scalability of such approximations, while achieving more reliable uncertainty quantification~\citep{he2026survey}, remains an important direction for AORL.

\textbf{Memory-based RL.} 
Memory remains underexplored in offline RL, with the most notable progress such as the Decision Transformer family~\citep{chen2021decision}, which still operate under conservative objectives. We believe that advances in understanding memory~\citep{ni2023transformers,cherepanov2026unraveling}, learning history representations~\citep{lambrechts2024informed,ni2024bridging,sinha2024agent}, and scaling memory-based RL~\citep{grigsby2024amago,luo2024efficient} in online POMDPs can help AORL by understanding how in-context learning emerges from offline-trained policies.\looseness=-1

\textbf{Beyond Bayesian model-based direction.} 
Although we emphasize the Bayesian model-based direction in \autoref{sec:bayesian} because it offers a principled interpretation of adaptive policy priors, simpler alternatives may also achieve adaptive behavior. These include model-free approaches~\citep{hu2024bayesian,wagenmaker2025posterior} as well as non-Bayesian approaches~\citep{touati2023does} (see \autoref{sec:related} for discussion). Understanding when such methods recover similar adaptive mechanisms, and when they fundamentally differ from model-based formulations, remains an important open direction.

\textbf{Foundation models for AORL.} World foundation models offer an intriguing answer to the \textit{prior} specification problem in \autoref{sec:bayesian}: rather than constructing $\Pr(\mathcal{M})$ from scratch, one can specify the MDP prior from a foundation model pretrained on broad dynamics data. Pretrained history-dependent policies,  including large language models and vision-language-action models, may further serve as strong initial policies that efficiently explore beyond offline dataset support.

\textbf{AORL for foundation models.}
Recent progress in foundation models already highlights the importance of test-time adaptation and continual learning~\citep{wei2025plangenllms,snell2025scaling}. Yet offline post-training remains dominated by supervised fine-tuning (SFT)~\citep{ouyang2022training}, which may narrow behavioral support and hinder later test-time adaptation or online RL fine-tuning~\citep{zhang2026good}. Bayesian directions may therefore offer a useful path for AORL in foundation models: model-based formulations can preserve broader support through synthetic on-policy rollouts~\citep{chen2025retaining}, while Bayesian formulations can promote in-context adaptation~\citep{zhang2026beyond}.

\section{Related Work and Research Areas}
\label{sec:related}

\textbf{Relation to prior adaptive offline RL work.}
Our position builds on prior work that develops algorithms for adaptive offline RL~\citep{ghosh2022offline,chen2021offline,choi2024diversification,ni2026long}. We do not claim that memory, exploration, or self-correction are individually new. Rather, our contribution is to identify their interaction as the capability structure of adaptability in offline RL and to elevate adaptability into an explicit objective of the offline stage. Prior work studies particular algorithmic realizations, primarily through Bayesian in-context adaptation, without organizing the broader problem around these three capabilities. AORL broadens adaptation beyond in-context learning to include planning, test-time training, and online fine-tuning, and uses this framing to motivate multi-episode evaluation and a broader research agenda.

\textbf{Plasticity in continual RL.} The ability to improve from subsequent experience is closely related to \textit{plasticity} in continual RL~\citep{abbas2023loss,klein2024plasticity} and offline-to-online RL~\citep{li2025three}. AORL differs in that adaptation often occurs without parameter updates at deployment time, also known as \textit{in-context plasticity}~\citep{klissarov2026improving}. Still, the central concern is similar: whether prior learning preserves the capacity to improve when new experience becomes available.

\textbf{Meta-RL.} The Bayesian perspective in \autoref{sec:bayesian} is closely related to contextual MDPs~\citep{hallak2015contextual} and meta-RL~\citep{duan2016rl,wang2016learning,beck2023survey}, where policies are trained across a distribution of tasks so that adaptation emerges at test time. The key difference is that in meta-RL the task distribution is pre-specified before training, whereas in Bayesian offline RL the distribution over environments must be self-constructed from offline data. Nevertheless, both settings give rise to similar adaptive behavior at test time, including online exploration and self-correction under uncertainty.

\textbf{Behavior foundation models.} 
A recent line of work trains behavior foundation models (BFMs) from \textit{reward-free} offline datasets, aiming to produce policies that generalize zero-shot to arbitrary downstream reward functions~\citep{touati2023does,park2024foundation,frans2024unsupervised}. The key idea is to learn representations that decouple dynamics from reward, so that a new task is solved by simple computation on a few externally-provided reward-labeled samples without further training. Once the task embedding is inferred, the policy is static and memoryless; dynamics are typically assumed fixed.\looseness=-1

BFMs are therefore complementary to AORL. Their reward generalization could help adaptive policy priors address test-time reward specification, but two gaps remain. First, adaptation in BFMs is typically passive, whereas AORL requires the agent to actively collect informative experience to identify the task. Second, AORL must also handle uncertainty in dynamics, particularly beyond offline support.

\vspace{-0.5em}
\section{Conclusion}
\label{sec:conclusion}
\vspace{-0.5em}

This paper argues that when policies are expected to continue improving through interaction, offline RL should prioritize learning \textit{adaptive policy priors} rather than static policies. Under this view, the objective of offline learning is not only immediate robustness under limited support, but also preserving the capacity for memory, exploration, and self-correction once new experience becomes available. Bayesian offline RL provides one principled direction for this perspective, though many algorithmic and theoretical questions remain open.
More broadly, we believe offline RL should gradually move beyond a supervised-learning view shaped by data scaling laws~\citep{kaplan2020scaling}, toward a continual-learning view aligned with the emerging era of experience~\citep{silver2025welcome} and learning-as-adaptation~\citep{abel2024three}.

\begin{ack}

We thank Yihao Sun and Dhruv Sreenivas for valuable comments on the manuscript. We thank Csaba Szepesvári for a stimulating discussion on the definition of offline RL and online RL, which partly motivated this paper. We also thank Hao Sun for introducing the notion of in-context plasticity. 

This work was funded by Canada CIFAR AI Chairs program.

\end{ack}

{
\small
\bibliography{main}
\bibliographystyle{plainnat}
}

\end{document}